\documentclass{article}
\usepackage{spconf,amsmath,graphicx}
\usepackage{amsfonts}
\usepackage{verbatim, tcolorbox}
\usepackage{multicol}
\usepackage{multirow}
\usepackage{graphicx}
\usepackage{hyperref}
\title{Confidence Estimation for LLM-Based Dialogue State Tracking}
\name{Yi-Jyun Sun, Suvodip Dey, Dilek Hakkani-Tür, Gokhan Tur }
\address{University of Illinois Urbana-Champaign}

\begin{document}
\ninept
\maketitle
\begin{abstract}
Estimation of a model's confidence on its outputs is critical for Conversational AI systems based on large language models (LLMs), especially for reducing hallucination and preventing over-reliance. In this work, we provide an exhaustive exploration of methods, including approaches proposed for open- and closed-weight LLMs, aimed at quantifying and leveraging model uncertainty to improve the reliability of LLM-generated responses, specifically focusing on dialogue state tracking (DST) in task-oriented dialogue systems (TODS). Regardless of the model type, well-calibrated confidence scores are essential to handle uncertainties, thereby improving model performance. We evaluate four methods for estimating confidence scores based on softmax, raw token scores, verbalized confidences, and a combination of these methods, using the area under the curve (AUC) metric to assess calibration, with higher AUC indicating better calibration. We also enhance these with a self-probing mechanism, proposed for closed models. Furthermore, we assess these methods using an open-weight model fine-tuned for the task of DST, achieving superior joint goal accuracy (JGA).
Our findings also suggest that fine-tuning open-weight LLMs can result in enhanced AUC performance, indicating better confidence score calibration. 

\end{abstract}

\section{Introduction}
As the adoption of dialogue systems grows, a critical challenge has emerged: ensuring the reliability of the responses of these systems and preventing the generation of model responses that are inaccurate or fabricated. To mitigate this problem, recent studies \cite{xiong2023can, tian2023just, kuhn2023semantic, mielke2022reducing} have focused on measuring model uncertainty to quantify the reliability of their outputs.
Reliability in dialogue systems refers to the system’s ability to consistently understand user inputs, retrieve relevant results or information when needed, generate appropriate responses, and handle uncertainties or errors effectively. A promising approach to improving reliability is through the estimation of confidence scores, which aim to provide a quantitative measure of the system’s uncertainty in its outputs. By incorporating confidence scores, dialogue systems can better manage uncertainties, identify potential errors, and make more informed decisions. For instance, a system with high confidence in its response can proceed smoothly, while one with low confidence can seek clarification from the user or escalate the query to a human operator. This dynamic adjustment based on confidence scores not only enhances the system’s reliability but also improves its overall performance and user satisfaction.

\begin{figure}[t]
\includegraphics[width=0.95\linewidth]{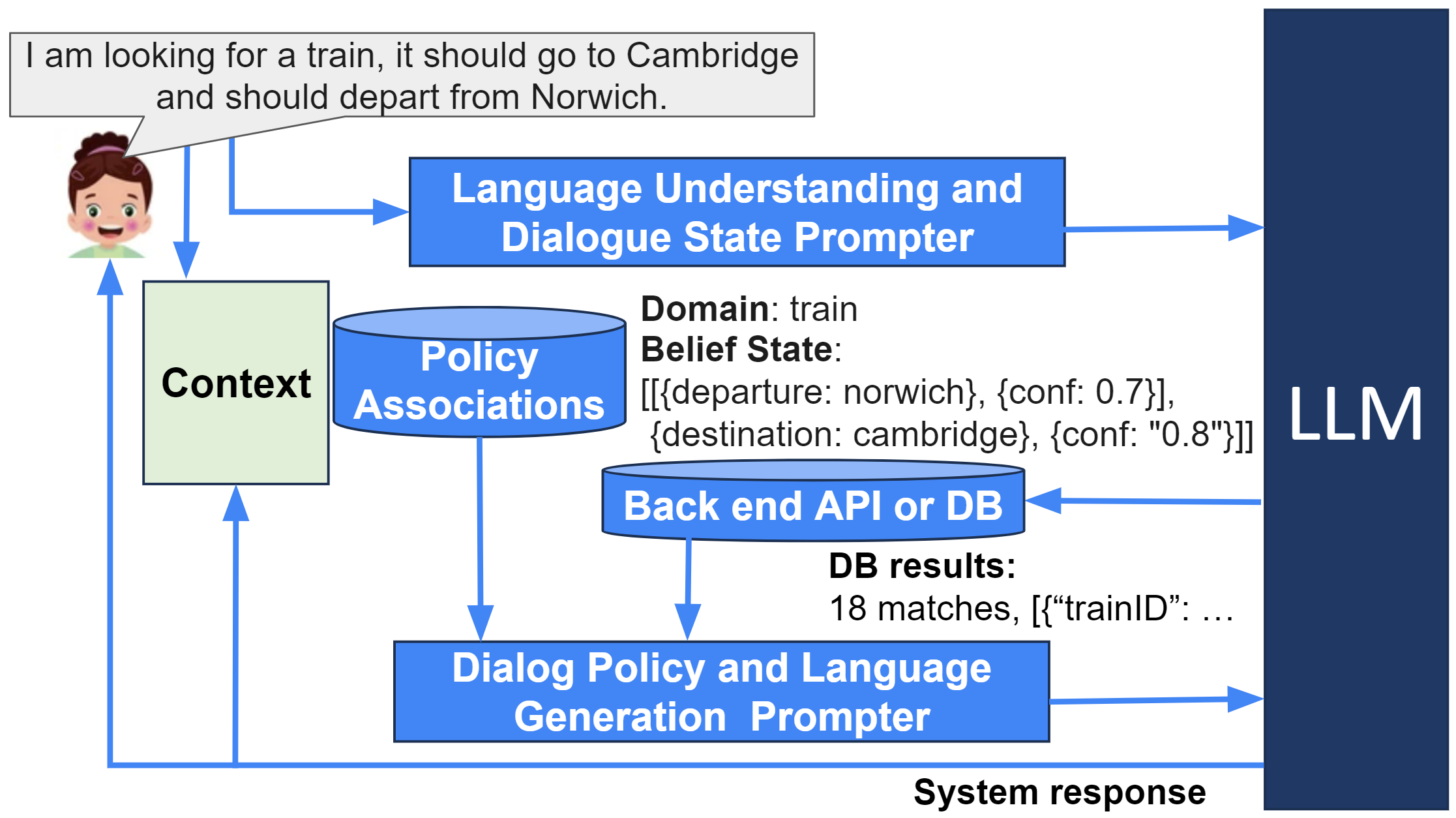}
\caption{Example interaction in our TODS approach, showing DST and its outputs with confidence scores.}
\label{dialog-example}
\end{figure}

To ensure that confidence scores can be applied reasonably, they must be well-calibrated. A well-calibrated confidence score means that the predicted probability accurately reflects the true likelihood of correctness, aligning the system's uncertainty with actual accuracy and making it possible to trust and utilize these scores. There are various methods for achieving well-calibrated confidence scores, including open-box and closed-box approaches that are proposed for open-weight and closed-weight models. Open-box approaches access internal model information such as model logits~\cite{duan2023shifting, kuhn2023semantic} and internal states~\cite{ren2022out, kadavath2022language, baan2022stop, li2024inference}, making them feasible only for open-weight models, such as Meta's Llama~\cite{touvron2023llama}. Closed-box methods, on the other hand, measure confidence on model outputs using verbalized or linguistic cues~\cite{mielke2022reducing, xiong2023can}, directly instructing a closed-weight model, such as OpenAI GPT-4~\cite{achiam2023gpt}, to express its confidence.

In this work, we focus on LLM-based DST for TODs. By instructing the LLM with slot descriptions and in-context examples, we predict slot-value pairs for an ongoing interaction. Instead of predicting complete outputs, we generate scores for individual slot-value pairs, offering finer-grained information for the dialoo policy as shown in Fig.~\ref{dialog-example}. This allows the system to confirm only uncertain slots with the user, reducing redundancy.

We explored four confidence estimation methods, applicable to both open-box and closed-box models, ensuring adaptability across various TODs architectures. Additionally, a self-probing prompting strategy improves the calibration of confidence scores, enabling systems to handle uncertainties more effectively. This approach lays the groundwork for more reliable dialogue systems that leverage enhanced confidence estimation.

The contributions of our work can be summarized as follows:
\begin{itemize}
\itemsep -0.5ex
\item We investigate various prompting strategies for dialogue state tracking and discuss their run time requirements and accuracy. 
\item We experiment with a set of open-box and closed-box methods for confidence estimation of dialogue state tracking.
\item We demonstrate that a combination of open-box and closed-box methods with open-weight LLMs results in the most reliable outcome.~\footnote{Code is available at \href{https://github.com/jennycs0830/Confidence_Score_DST}{github.com/jennycs0830/Confidence\_Score\_DST}}
\item We show that applying the self-probing strategy can improve the calibration level of confidence scores. 
\end{itemize}

\section{Related Work}
\subsection{Dialogue State Tracking}
Dialogue State Tracking is crucial in TODs, aiming to capture the user's goals and intent during conversations. DST takes the user's utterance and conversation history as input and outputs the dialogue belief state in a slot-value format, based on a domain-specific schema. This belief state guides the system's next action, as shown in Fig.~\ref{dialog-example}.

The nature of DST requires a predefined domain-specific schema, and training a DST model necessitates annotated domain-specific dialogues as training data. However, data collection is notoriously challenging and labor-intensive. Consequently, the ability to handle unseen domains in a zero-shot manner becomes a crucial capability for DST systems~\cite{lin2021zero, lin2021leveraging, campagna2020zero}.

\subsection{Model Uncertainty}
In machine learning, there are two types of uncertainty: epistemic~\cite{hullermeier2021aleatoric} and aleatoric uncertainty. Epistemic uncertainty arises from limited knowledge and reflects uncertainty in the model parameters. By measuring it, one can assess the model's reliability—lower epistemic uncertainty typically indicates more accurate predictions with higher confidence. In contrast, aleatoric uncertainty stems from inherent randomness or noise in the data, representing the variability within the data itself.

To quantify these uncertainties effectively, we categorize existing uncertainty measurement methods into two categories. The first is open-box methods, which access the model's internal information. Logits-based approaches, like~\cite{duan2023shifting, kuhn2023semantic}, estimate model uncertainty by utilizing model weights during inference. Ensemble-based approaches, like~\cite{manakul2023selfcheckgpt, lin2023generating, van2023camell}, estimate model uncertainty by calculating the consistency between multiple responses or extracting from the predictive distribution of those ensemble models. Furthermore, \cite{van2023camell} trains a single model by adopting ensemble distillation. The second category is methods for closed models, which do not access the internal model. Instead, these methods estimate uncertainty mostly by prompting the model~\cite{kojima2022large, wang2023plan, tian2023just, xiong2023can} in different strategies and using the responses as cues to estimate model uncertainty. 

Our method differs from traditional ensemble-based approaches by combining generation, non-generation, and verbalized insights to estimate uncertainty using a single LLM output, rather than multiple model runs. Although slot-level self-probing increases computational cost slightly, our approach remains more efficient than ensembles, which require multiple predictions. This makes our method both computationally efficient and broadly applicable, offering robust uncertainty estimation without the heavy overhead of ensemble methods.
% In our work, we propose a method that combines generation, non-generation, and verbalized insights to achieve more informative confidence scores. Our approach not only avoids the computational cost associated with ensemble-based methods but also incorporates a broader range of information beyond just model logits.

\subsection{Confidence Calibration}
Recent studies have explored various methods to produce well-calibrated confidence scores. Authors of~\cite{zhao2021calibrate} employ mathematical methods to calibrate the model's output probabilities. It quantifies the model's bias towards certain answers using content-free input, adjusting the probabilities to better reflect true confidence levels. Another promising approach is the use of prompting strategies~\cite{tian2023just}. This study found that verbalized confidence, where the model expresses its confidence in natural language, is typically better calibrated than the model's raw conditional probabilities.

\section{Approach}

In order to present confidence estimation approaches, we first present our dialogue state tracking method inspired from the recent state-of-the-art work. 

\subsection{Dialogue State Tracking}
Our dialogue state tracking approach is based on the LLM-based method proposed in~\cite{hudecek-dusek-2023-large}. Basically, an LLM is prompted twice for each turn, once for detecting the current domain (such as restaurant or train) and then the follow-up for slot filling, i.e., assigning slot values (such as restaurant name or cuisine) for that turn for the selected domain. This strategy greatly reduces the number of candidate slots and allows writing more targeted prompts. The prompts contain a description of the task along with examples. All prompts used in this paper are provided as supplementary material. A simplified version of the prompt used for domain classification is shown below.

\scriptsize
\begin{verbatim}
Determine which domain is considered in the
following dialogue situation:
- restaurant
- hotel
...
[Examples]
...
\end{verbatim}

\normalsize
\noindent
The dialogue history is given as input to the LLM. Since the model is focused on the current turn, slot carryover is performed manually following the MinTL approach~\cite{lin-etal-2020-mintl}, as done in the previous work. This is especially critical for the dialogues spanning multiple domains, such as first finding an attraction place followed by booking a taxi to go there. 

We explored multiple prompt designs, varying along two dimensions: zero-shot vs. few-shot and predicting all slots at once vs. one slot at a time. Few-shot prompts include DST-related examples, while asking for all slots allows a single LLM call, compared to multiple calls for individual slots. These four prompt variants are detailed in the supplementary material. Like previous work, we instructed the LLM to output in JSON format, using in-context examples. A simplified slot-filling prompt for a single slot is shown below.

\scriptsize
\begin{verbatim}
In the domain of ``train'', extract ``departure'' 
slot which specifies the departure station
Do not capture any other values. If not 
specified,  do not respond to that slot-value.
---------------------Example:
context: Customer: I need to get a train from 
cambridge to Norwich.
belief state: {departure: cambridge}
\end{verbatim}

\normalsize

\subsection{Confidence Score Estimation}

In this study, we experiment with both open- and closed-weight models. Open-weight models serve two purposes, (i) comparing the confidence measurements directly estimated from the model with the closed-weight model approaches, including self probing, and (ii) assessing the effect of task specific fine-tuning on confidence estimation. However, note that, as the current open-weight models are less powerful than the closed-weight models, we report both DST accuracy related metrics and confidence score metrics. Our goal is devising an approach that maximizes metrics for both DST accuracy and confidence score quality.

\begin{figure}[t]
\includegraphics[width=0.95\linewidth]{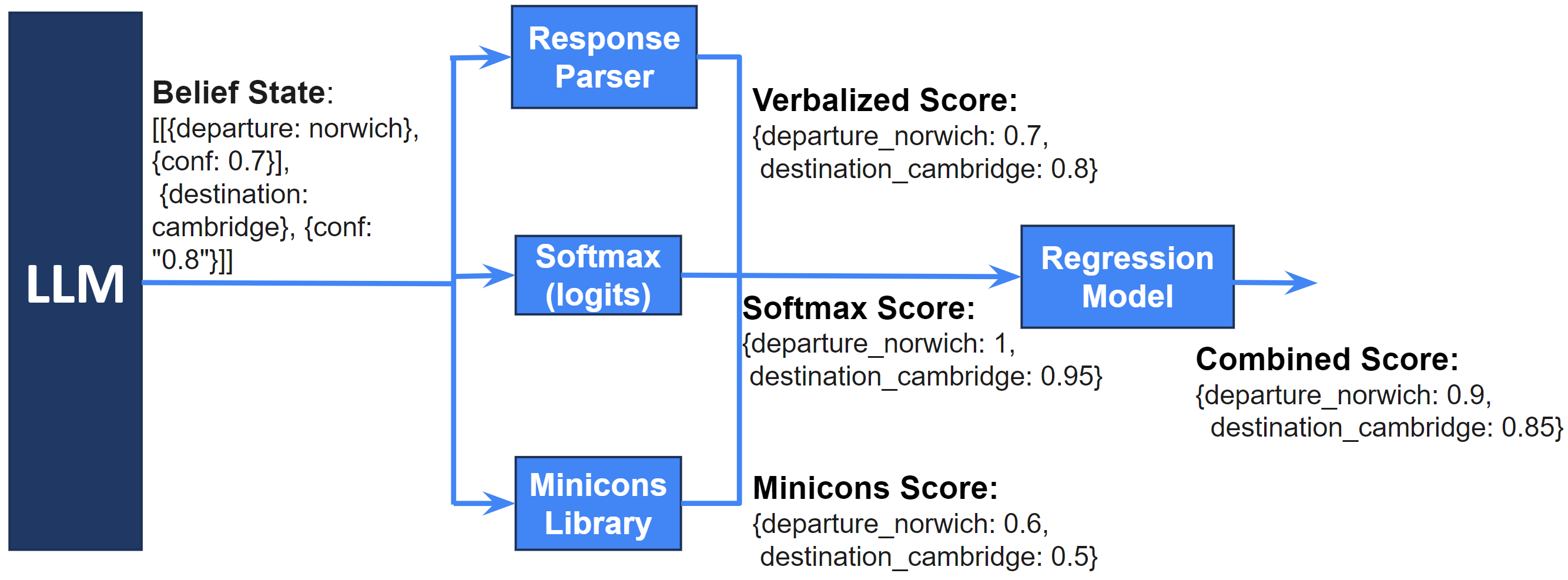}
\caption{An example demonstrating the individual and combined confidence scores.}
\label{conf-example}
\vspace*{-1ex}
\end{figure}

\subsubsection{Methods for Open-Weight Models} We adopt two methods to estimate the confidence score for open-weight models, described as follows.

\textit{A. Scores derived from Token Probabilities of the LLM:} Given the model in a generative fashion, the logits and softmax outputs reflect the token probability distribution of the model conditioned on dialogue context $C$. Logits ($\vec{l}$), are crucial because they contain the raw prediction information, which can be transformed into confidences by applying the softmax activation function as,
\begin{equation}
    \mathrm{Conf}_i = \sigma(\vec{l})_i = \frac{\exp(l_i)}{\sum_{j} \exp(l_j)} 
\label{eq: softmax logits}
\end{equation}
where $\mathrm{Conf}_i$ and $l_i$ denotes the confidence and logits of token $i$, respectively. Beam search then uses these probability-form logits to calculate the scores for the sequences and extends the current candidate sequences to generate new candidate sequences as follows.
\begin{equation}
S(C \oplus i) = S(C) + \log P(i\mid C)    
\label{eq: beam search}
\end{equation}
where $S(C \oplus i)$ represents the score for the concatenation of the context and token $i$. We calculate it by adding the score of the existing sequence/context $S(C)$ to the log-probability of the new token $i$ given the existing sequence, i.e., $\log P(i \mid C)$.

In practice, a single word of the slot name or value may be composed of multiple tokens. To obtain the confidence score for a word with $N$ tokens, we combine the logits ($l_i$) of each token in the word sequence, defined as follows,

\begin{equation}
   \mathrm{Conf}_w = \prod_{j=1}^{N} \sigma(\vec{l})_j
\label{eq: word confidence}
\end{equation}
where $\mathrm{Conf}_{w}$ represents the confidence score of the word $w$ with $N$ tokens. This formula calculates the word confidence by taking the product of the softmax probabilities of each token, effectively combining the individual token confidences into a single measure of confidence for the entire word.

Additionally, the confidence score for a slot-value pair is derived from the individual confidence of the slot and the value tokens. We denote it as $\mathrm{Conf}_{\text{slot-value}}^{(\text{Softmax})}$ as it is computed using the softmax scores. This score is calculated as follows:

\begin{equation}
\mathrm{Conf}_{\text{slot-value}}^{(\text{Softmax})} =  \mathrm{Conf}_{\text{slot}} \times \mathrm{Conf}_{\text{value}}
\label{eqn:conf_slot_value}
\end{equation}
where $\mathrm{Conf}_{\text{slot}}$ and $\mathrm{Conf}_{\text{value}}$ represent the confidence scores of the slot and the value, respectively. By multiplying these two confidence scores, we obtain a pair confidence score that reflects the confidence in the accuracy of both the slot and its corresponding value. In our preliminary experimentation, we tested whether using just the score of the slot or the value may result in better performance for the slot-value pair and converged to using the multiplied confidence in the equation above. From the task completion perspective, both the slot and the value should be accurate for retrieving accurate information from knowledge sources.\\

\textit{B. Scores derived using The Surface Form:} This is a complementary approach, showing the logits in a non-generative fashion, so no beam search is necessary. We just compute the score of each given token in a response string using the LLM:

\begin{equation}
\mathrm{Conf}_{i}^{(\text{LLM})} = \log P(i|C_{<i};\theta)   
\end{equation}
where $\mathrm{Conf}_{i}^{(\text{LLM})}$ denotes the confidence score of token $i$ with context $C_{<i}$, using the LLM with weights $\theta$ following the Llama/GPT-style architecture~\cite{radford2019language}. We have used the minicons library\footnote{https://pypi.org/project/minicons/} to compute these raw token-level scores. For slot-value pairs with multiple tokens, we aggregated the token-level scores by taking an average. For a given slot-value pair, we denote the minicons confidence score $\mathrm{Conf}_{\text{slot-value}}^{(\text{Minicons})}$ as follows,
\begin{equation}
    \mathrm{Conf}_{\text{slot-value}}^{(\text{Minicons})} = \frac{1}{N} \sum_{j=1}^{N} \mathrm{Conf}_{j}^{(\text{LLM})}
\end{equation}
where the slot-value pair is composed of $N$ tokens.

\subsubsection{Methods for Open- and Closed-Weight Models}

The following two methods can be applied to both open- and closed-weight models. These are also helpful in comparing their performances for confidence estimation. Furthermore, an open model, like Llama, fine-tuned with the target DST task, can also be used with these approaches.

% \paragraph{Verbalized Confidence Estimation}
\textit{A. Verbalized Confidence Estimation:} In addition to utilizing open-weight models, we also implement a concept inspired by human conversation. We prompt the model to adhere to a predefined output format and directly extract the verbalized confidence score for each slot-value pair from the text generated by the LLM. For a given slot-value pair, this score is denoted as $\mathrm{Conf}_{\text{slot-value}}^{(\text{Verbalized})}$. This approach leverages the model's ability to articulate its confidence levels in a human-like manner, providing a more transparent means of assessing the reliability of its predictions. A simplified prompt used for this approach is shown as follows.

\scriptsize
\begin{verbatim}
Capture entity values with confidences given 
this  conversation,  focusing on only the 
values  mentioned in the  last utterance.
Output Format: 
{state: {_entity_:_value_}, conf: X}
In the domain of "train", the values that 
should be captured are:
- "arriveby" that specifies what time the 
train should arrive
...
\end{verbatim}
\normalsize

% \paragraph{Self-Probing Confidence Estimation}
\textit{B. Self-Probing Confidence Estimation:} Humans frequently notice that it is often simpler to spot errors in others' responses than in their own. Inspired by the approach in \cite{xiong2023can}, where the model is queried with ``\textit{how likely is the above answer to be correct?}'' to investigate model uncertainty improvements, we integrated this self-probing prompting strategy into our confidence estimation process. Specifically, we employed self-probing in three distinct scenarios: i) no self-probing, ii) self-probing at the turn-level, and iii) self-probing at the slot-level.

In our method, after predicting the belief state, we provided the predicted response from the previous state to the self-probing prompt. We then asked the model to analyze the input belief state, generate a confidence score ranging from 0 to 1, and give a brief reason for the prediction. This process helps in evaluating the model's uncertainty and improving the overall accuracy of the predictions. The key difference in our approach is the granularity at which self-probing is applied, offering insights at both the turn and slot levels, thereby enhancing the robustness of the confidence estimation. For the slot-level self-probing, we made a probing call to the LLM for every slot value in the state. On the other hand, in the turn-level self-probing, we provide all slot-value pairs in the belief state and ask the model to provide scores for each of them. Note that the slot-level self-probing is more expensive as it requires several LLM inferences. A simplified prompt for turn-level self-probing is shown below.
\scriptsize
\begin{verbatim}
How likely is the below state to be correct?
Analyze the state, provide a brief reason, and
give confidence (0-1).
’Customer: I need train reservations from 
norwich to cambridge’
State: {’departure’: ’norwich’, ’destination’: 
’cambridge’, ’bookpeople’: ’2’}
\end{verbatim}

\normalsize

\subsubsection{Combined Confidence Estimation}
In the previous sections, we introduced three types of confidence score: i) softmax confidence $\mathrm{Conf}_{\text{slot-value}}^{(\text{Softmax})}$, ii) minicons confidence $\mathrm{Conf}_{\text{slot-value}}^{(\text{Minicons})}$, and iii) verbalized confidence $\mathrm{Conf}_{\text{slot-value}}^{(\text{Verbalized})}$. The first two scores are applicable to open-weighted models, while the third one works for both open and closed-weighted models. The reason for estimating multiple types of confidence scores is to provide more insights and generate calibrated confidence scores. Fig.~\ref{conf-example} depicts an example of the confidence score combination for DST. 

To achieve this, for each experimental setting with open models, we train a linear regression model to produce a combined confidence score for a given slot value pair as follows:

\begin{equation}
    \mathrm{Conf}_{\text{slot-value}}^{(\text{Combined})} = \alpha \mathrm{Conf}_{\text{slot-value}}^{(\text{Softmax})} + \beta \mathrm{Conf}_{\text{slot-value}}^{(\text{Minicons})} + \gamma \mathrm{Conf}_{\text{slot-value}}^{(\text{Verbalized})}
\end{equation}
where $\alpha,\beta, \gamma \in \mathbb{R}$ are the learnable parameters of the linear regression model and denote the weightage for each confidence score.

\begin{table}[t]
\centering
\begin{tabular}{lrrr}
\hline \textbf{Data} & \textbf{\#Dialogues} & \textbf{\#Turns} & \textbf{Avg turns per dialogue}  \\ \hline
Train & 8420 & 56668 & 6.73 \\
Dev & 1000 & 7374 & 7.37 \\
Test & 999 & 7368 & 7.37 \\
\hline

\end{tabular}
\caption{\label{tbl:stat-Mul} Data statistics of MultiWOZ 2.2 dataset
}
\vspace{-0.1in}
\end{table}

\section{Experiment Setup}
\subsection{Dataset}
Our experiments use MultiWOZ~\cite{budzianowski}, a multi-domain task-oriented dialogue dataset. It is a human-human written dialogue dataset that contains turn-level annotations and descriptions of slot labels. We use the MultiWOZ 2.2~~\cite{multiwoz22} version, which has refined belief state and user action annotations. We use the training subset of MultiWOZ to fine-tune the open-weight models and the validation set to train regression for combining the confidence scores from multiple methods. The basic statistics of  the dataset are shown in Table~\ref{tbl:stat-Mul}.

\subsection{Models}
We evaluate our method on three models: the closed-weight GPT-4, the open-weight Llama3-8B, and the open-weight Mistral-7B. For both Llama3-8B and Mistral-7B, we also use versions that are fine-tuned with the training sets of both the MultiWOZ and SGD datasets. Additionally, 8-bit quantization is applied to both models to optimize performance and reduce memory usage. 

\subsubsection{Fine-tuning Details}
To enable the model to generate responses that include verbalized confidence scores, we built ground truth confidence scores for use during the fine-tuning phase. For each dialogue turn, the model assesses the difficulty of predicting the dialogue state based on the given user utterance and dialogue history. Below is the simplified prompt we have used to assess the difficulty:

\scriptsize
\begin{verbatim}
How difficult would it be for a Language Model 
to predict the dialogue state from:
[utterance]
given dialogue history
[history]

Choose the level of hardness from (Easy/Medium/
Hard).
Answer:
\end{verbatim}

\normalsize
In this work, we use four difficulty levels - High, Medium, Easy, and Other. The difficulty level is mapped into a confidence score by introducing a degree of randomness appropriate to each level, described as follows.

\small
\begin{itemize}
    \item Easy: mapped to a range between 0.9 and 1.0.
    \item Medium: mapped to a range between 0.8 and 0.9.
    \item Hard: mapped to a range between 0.7 and 0.8.
    \item Other: A default confidence score of 0.5 is assigned.
\end{itemize}

\normalsize
This mapping process adds variability to the confidence scores, better reflecting real-world uncertainty. During the fine-tuning process, we provide the model with both the ground truth state and the corresponding ground truth confidence score for each slot-value pair. This dual-training approach aims to enhance the model's ability to accurately predict dialogue states and verbalize its confidence, thereby improving the overall performance and reliability of the dialogue state tracking system.

During fine-tuning, we provide the model with both the ground truth state and the corresponding ground truth confidence score for each turn. This enhances the model's ability to predict verbalized confidence and improves the overall performance. Additionally, we employed low-rank adapters (LoRA)~\cite{lora} to fine-tune the model.

\subsection{Evaluation Metrics}
We evaluated our method in two aspects: i) the performance of our approach on the DST task and ii) the calibration level of the confidence scores.

\subsubsection{DST Evaluation Metrics}
To evaluate the quality of dialogue state tracking, we use two metrics: joint goal accuracy (JGA) and slot-level F-measure (Slot-F). JGA is the commonly used DST evaluation metric that requires all the slot and value pairs in the predicted state to exactly match the slot and value pairs in the ground truth state, for that turn to be considered accurate. Previous work discussed that JGA overly punishes DST performance~\cite{dey-etal-2022-towards, kim-etal-2022-mismatch}. Hence, we also present slot F-measure, which computes the slot-value pair level F-scores between the predicted and ground truth dialogue states. 

\subsubsection{Confidence Score Evaluation Metrics}

To evaluate the calibration of confidence scores, we use two metrics: ROC-AUC (Receiver Operating Characteristic Area Under Curve ) and ECE (Expected Calibration Error). ROC-AUC measures the model's ability to distinguish between correct and incorrect slot-value pair predictions based on confidence scores. A higher AUC indicates better discrimination between correct and incorrect predictions. ECE assesses how well the confidence scores are calibrated by comparing predicted confidence levels with actual accuracy across different confidence bins. A lower ECE score means better calibration, as it reflects smaller differences between confidence and actual accuracy.

\section{Results}
In our experiments, we use the MultiWOZ 2.2 test data for assessing the quality of the dialogue state tracking and associated confidence score estimations.

\subsection{Dialogue State Tracking Performance}

Table~\ref{tab:dst} shows the JGA performance of various prompting strategies for DST. We use four prompting strategies using descriptions of All/One slot(s) with examples covering All/One slot(s). For example, in the All/All strategy, we provide descriptions of all slots with examples covering all slots. In the All/One strategy, we provide descriptions of all slots with the example covering a single slot. The other two strategies can be defined in a similar fashion.

We observe that the All/All strategy achieves the best JGA score for different models. 
This is because it covers all the slots along with examples, resulting in superior performance. Moreover, the All/All strategy is also the most computationally efficient method. As it covers both descriptions and examples of all the slots, this strategy requires calling the LLM only once for every dialogue turn. In contrast, the other three strategies necessitate multiple calls to the LLM, significantly increasing the computational overhead. Given the advantages in both performance and efficiency, we adopt the All/All strategy to report the results for the remainder of the paper.

\begin{table}[t]
\begin{center}
\begin{tabular}{c|r|r}
\hline
Strategy&GPT-4&TK-Instruct-11B\\
\hline
All/All&{\bf 40.8\%}&{\bf 34.9\%}\\
All/One&31.8\%&8\%\\
One/All&37.3\%&32.6\%\\
One/One&32.9\%&13.9\%\\
\hline
\end{tabular}
\end{center}
\caption{Performance of DST in terms of JGA for various prompting strategies using descriptions of All/One slot(s) with examples covering All/One slot(s).}
\label{tab:dst}
\vspace{-0.1in}
\end{table}

\begin{table*}[t]
    \centering
    \begin{footnotesize}
    \begin{minipage}{0.48\linewidth}
        \centering
        \begin{tabular}{|c|c|c|c|c|c|c|}
        \hline
        \multicolumn{7}{|c|}{\textbf{Zero-shot}} \\
        \hline
        Model & Self- & Confidence & \multicolumn{4}{|c|}{Results}\\
        \cline{4-7}
        & Probing & Estimation & JGA & Slot-F & AUC & ECE \\
        \hline
        Llama3 & no & softmax & & & 0.624 & 0.164 \\
        \cline{3-3} \cline{6-7}
         & & minicons & & & 0.514 & 0.265 \\
        \cline{3-3} \cline{6-7}
         & & verbalized & & & 0.561 & 0.364 \\
         \cline{3-3} \cline{6-7}
         & & combined & & & 0.655 & 0.032 \\
         \cline{2-3} \cline{6-7}
         
         & turn & softmax & & & 0.645 & 0.166\\
        \cline{3-3} \cline{6-7}
         & & minicons & 14.7 & 68.7 & 0.533 & 0.265  \\
        \cline{3-3} \cline{6-7}
         & & verbalized & & & 0.597 & 0.205\\
         \cline{3-3} \cline{6-7}
         & & combined & & & 0.657 & 0.011\\
         \cline{2-3} \cline{6-7}

         & slot & softmax & & & 0.619 & 0.340 \\
        \cline{3-3} \cline{6-7}
         & & minicons & & & 0.520 & 0.238 \\
        \cline{3-3} \cline{6-7}
         & & verbalized & & & 0.566 & 0.351 \\
         \cline{3-3} \cline{6-7}
         & & combined & & & 0.656 & 0.035 \\ 
        \hline
        
        Finetuned & no & softmax & & & 0.722 & 0.164 \\
        \cline{3-3} \cline{6-7}
         Llama3& & minicons & & & 0.514 & 0.265 \\
        \cline{3-3} \cline{6-7}
         & & verbalized & & & 0.506 & 0.310 \\
         \cline{3-3} \cline{6-7}
         & & combined & & & {\bf 0.725} & {\bf 0.018} \\
         \cline{2-3} \cline{6-7}
         
         & turn & softmax & & & 0.682 & 0.113 \\
        \cline{3-3} \cline{6-7}
         & & minicons & 44.6 & 88.3 & 0.517 & 0.344 \\
        \cline{3-3} \cline{6-7}
         & & verbalized & & & 0.556 & 0.208 \\
         \cline{3-3} \cline{6-7}
         & & combined & & & 0.687 & 0.053\\
         \cline{2-3} \cline{6-7}
         
         & slot & softmax & & & 0.720 & 0.165 \\
        \cline{3-3} \cline{6-7}
         & & minicons & & & 0.506 & 0.205 \\
        \cline{3-3} \cline{6-7}
         & & verbalized & & & 0.514 & 0.305 \\
         \cline{3-3} \cline{6-7}
         & & combined & & & 0.724 & 0.021\\
         \cline{2-3} \cline{6-7}
        \hline

        GPT4 & no & &  &  & 0.529 & 0.345 \\
        \cline{2-2} \cline{6-7}
          & turn & verbalized & 36.1 & 82.7 & 0.542 & 0.303 \\
         \cline{2-2} \cline{6-7}
         & slot & & & & 0.530 & 0.332 \\
         \hline
    \end{tabular}
    \end{minipage}
    \hfill
    \begin{minipage}{0.48\linewidth}
        \centering
        \begin{tabular}{|c|c|c|c|c|c|c|c|}
        \hline
        \multicolumn{7}{|c|}{\textbf{Few-shot}} \\
        \hline
        Model& Self- & Confidence & \multicolumn{4}{|c|}{Results}\\
        \cline{4-7}
        & Probing & Estimation & JGA & Slot-F & AUC & ECE \\
        \hline
        Llama3  & no & softmax & & & 0.603 & 0.258 \\
        \cline{3-3} \cline{6-7}
         & & minicons & & &  0.521 & 0.264  \\
        \cline{3-3} \cline{6-7}
         & & verbalized & & & 0.563 & 0.343 \\
         \cline{3-3} \cline{6-7}
         & & combined & & & 0.635 & 0.021 \\
         \cline{2-3} \cline{6-7} 
         
         & turn & softmax & & & 0.617 & 0.124 \\
        \cline{3-3} \cline{6-7}
         & & minicons & 26.5 & 75.7& 0.536 & 0.317\\
        \cline{3-3} \cline{6-7}
         & & verbalized & & & 0.608 & 0.162 \\
         \cline{3-3} \cline{6-7}
         & & combined & & & 0.644 & 0.020 \\
         \cline{2-3} \cline{6-7}
         
         & slot & softmax & & & 0.578 & 0.299 \\
        \cline{3-3} \cline{6-7}
         & & minicons & & & 0.519 & 0.244 \\
        \cline{3-3} \cline{6-7}
         & & verbalized & & & 0.566 & 0.379 \\
         \cline{3-3} \cline{6-7}
         & & combined & & & 0.621 & 0.028 \\
         \cline{2-3} \cline{6-7}
        \hline
        
        Finetuned & no & softmax &  & & 0.749 & 0.168 \\
        \cline{3-3} \cline{6-7}
         Llama3 & & minicons & & & 0.527 & 0.259 \\
        \cline{3-3} \cline{6-7}
         & & verbalized & & & 0.506 & 0.278 \\
         \cline{3-3} \cline{6-7}
         & & combined & & & 0.752 & 0.016 \\
         \cline{2-3} \cline{6-7}
         
         & turn & softmax &  & & 0.709 & 0.126 \\
        \cline{3-3} \cline{6-7}
         & & minicons & 35.8 & 84.4 & 0.532 & 0.340 \\
        \cline{3-3} \cline{6-7}
         & & verbalized & & & 0.539 & 0.189 \\
         \cline{3-3} \cline{6-7}
         & & combined & & & 0.715 & 0.057 \\
        \cline{2-3} \cline{6-7} 
         
         & slot & softmax &  & & 0.763 & 0.167 \\
        \cline{3-3} \cline{6-7}
         & & minicons & & & 0.536 & 0.257 \\
        \cline{3-3} \cline{6-7}
         & & verbalized & & & 0.504 & 0.292 \\
         \cline{3-3} \cline{6-7}
         & & combined & & & {\bf 0.766} & {\bf 0.018} \\
        \hline

        GPT4 & no & & & & 0.522 & 0.355\\
        \cline{2-2} \cline{6-7}
          & turn & verbalized & 40.8 & 84.7 & 0.541 & 0.324\\
         \cline{2-2} \cline{6-7}
         & slot & & & & 0.528 & 0.343 \\
         \hline
        \end{tabular}
    \end{minipage}
    \end{footnotesize}
    \caption{Experimental zero-shot (left) an few-shot (right) results for dialogue state tracking performance (JGA and Slot-F) and confidence estimation quality (AUC and ECE).}
    \label{tab:alternative}

\vspace{-0.1in}
\end{table*}

Table~\ref{tab:alternative} presents the results for the dialogue state tracking and confidence score prediction experiments using two models - i) closed-weight GPT-4 and ii) open-weight Llama3-8B. The left table presents results with no DST examples in the instruction context, whereas the right table presents results with few-shot examples included in the context of the instruction to the model. For few-shot, we used three examples in these experiments. The examples are selected by utilizing FAISS DB\footnote{https://github.com/facebookresearch/faiss} to find similar samples to the dialogue that is being tested from the training subset based on the last two utterances. We also present results after fine-tuning with the training subset of the MultiWOZ 2.2. 

In Table~\ref{tab:alternative}, we can observe that for the zero-shot scenario with no in-context examples in the instruction, the GPT-4 outperforms the non-finetuned Llama3. However, the finetuned Llama3 (trained using the MultiWOZ training set) results in the highest JGA of 44.6\%, outperforming the GPT-4 model. This shows that handling uncertainty using confidence scores can boost the performance of zero-shot DST prediction. On the other hand, for the few-shot scenario, GPT-4 outperforms the Llama3 models. 
% \suvodip{add explanation ....}. 
However, we can observe two additional things here. Firstly, The performance of the fine-tuned Llama3 is not far behind GPT-4, which is encouraging. Secondly, it demonstrates a superior confidence score, explained in detail in Section~\ref{sec:confidence}.

Additionally, experiments using the Mistral-7B model, detailed in Appendix Section G, show similar trends, further validating the applicability of our methods to different open-weight models.

% with no in-context examples in the instruction input to the model. This performance surpasses that of the GPT4 with in-context examples sampled from the training dataset.

% the JGA score with Llama-3 is about 0.14, whereas GPT-4 results in JGA of about 0.35 using a zero-shot approach. Fine-tuning Llama-3 using the MultiWOZ training set results in the highest JGA of 0.446 with no in-context examples in the instruction input to the model. This performance surpasses that of the GPT4 with in-context examples sampled from the training dataset.

\subsection{Quality of the Confidence Scores}
\label{sec:confidence}
Besides JGA, Table~\ref{tab:alternative} also presents the performance of confidence scores with respect to ROC-AUC (shown as AUC in Table~\ref{tab:alternative}) and ECE. While the JGA of GPT-4 with in-context examples (few-shot) is high, the quality of the verbalized confidence scores is lower than other models, even with self-probing. The best AUC obtained with GPT-4 is around 0.54 for both zero-shot and few-shot scenarios. Amongst the individual methods, the softmax confidence score achieves the best results for both metrics. Fine-tuning Llama3 for the task also results in an improvement of the confidence score quality, similar to the JGA performance. This result is aligned with the previous works~\cite{sicilia2024dealdealorknows}. The combination of the three confidence scores leads to small improvements for AUC. However, this improvement is much higher for ECE due to better calibration of the score values after the regression. We also observe performance gain after applying self-probing in some instances. For example, Llama3, with few-shot examples and a combined confidence score, achieves the highest AUC. In terms of the two targets of high DST accuracy and best confidence scores, we get the best outcome with the fine-tuned Llama3 model using the combined confidence scores, resulting in a JGA of 44.6\%, AUC of 0.725 and ECE of 0.018.

\subsection{Correlation Analysis}
We analyze the correlation of confidence scores using the Pearson correlation coefficient, based on the MultiWOZ dataset with Llama3 experiments. A label list is created, where 1 indicates the presence of a slot-value pair in the ground truth, and 0 otherwise. Table~\ref{tab:my_label} shows the correlations of the four confidence scores with these labels. Softmax scores show moderate correlation, while minicons and verbalized scores have weak correlations in both zero-shot and few-shot scenarios. This also justifies the superior performance of the combined confidence score in Table~\ref{tab:alternative}. Although the improvement in the correlation for the combined score is marginal with respect to the softmax confidence score, it results in better calibration.

\begin{table}[t]
    \centering
    \begin{footnotesize}
    \begin{tabular}{|c|c|c|c|c|c|c|c|}
    \hline
         \textbf{-shot} & \textbf{Self-} & \multicolumn{4}{|c|}{\textbf{Pearson coefficient}}\\
        \cline{3-6}
        & \textbf{Probing} & \textbf{Softmax} & \textbf{Minicons} & \textbf{Verbalized} & \textbf{Combined} \\
        \hline
         zero & no & 0.335 & 0.062 & 0.015 & 0.336 \\
        \cline{2-6}
         & turn &  0.278 &  0.082 & 0.185 &  0.285\\
        \cline{2-6}
         & slot & 0.251 & 0.039 & 0.211 & 0.311 \\
        \hline
         few & no & 0.440 & 0.046 & 0.083 & 0.443 \\
        \cline{2-6}
         & turn & 0.355 & 0.057 & 0.089 & 0.373 \\
        \cline{2-6}
         & slot &  0.466 & 0.060 & 0.056 & 0.467 \\
        \hline
    \end{tabular}
    \caption{Correlation between confidence score and correctness.}
    \label{tab:my_label}
    \end{footnotesize}
\vspace{-0.1in}
\end{table}

\subsection{Computational Costs}
While slot-level self-probing incurs higher computational costs due to the need for multiple LLM inferences per slot, our results show that turn-level self-probing, which requires only a single LLM inference per turn, offers comparable performance. As demonstrated in Table~\ref{tab:alternative}, the differences in evaluation metrics between slot-level and turn-level are minimal, with variations of less than 0.05. Therefore, turn-level self-probing provides a much more computationally efficient solution without sacrificing performance, making it a more scalable option for practical use.
\section{Conclusion and Future Work}
In summary, we presented a comprehensive study exploring various methods for confidence score estimation for dialogue state tracking (DST) in a Conversational AI system. In this work, we explored four different confidence scores based on softmax, raw token scores, verbalized confidences, and a combination of these methods. We observed that all methods are sub-optimal with closed-weight models, and it is possible to get well-calibrated confidence scores by fine-tuning an open-weight model. Regarding how to use the confidence score we get, existing work such as \cite{van-niekerk-etal-2021-uncertainty} incorporate uncertainty in the database query vector in the form of confidence thresholds, or \cite{van2023camell} use confidence score at data selection components and label-validation mechanism. In our work, we found that incorporating confidence scores into fine-tuning procedure significantly improves the DST performance. In addition, a well-calibrated confidence score is generated by the combined method. We also showed that the combined confidence score is moderately correlated with the ground truth labels, justifying its superior performance. 

Our future work involves using these confidence scores to improve the goal completion rate by extending the dialogue policy model accordingly. For example, if the system has low confidence about a slot value, it can perform implicit or explicit confirmation, or utilize this in a chain of thought reasoning process. Furthermore, we plan to conduct experiments on a second dataset like SGD~\footnote{Available at \href{https://github.com/google-research-datasets/dstc8-schema-guided-dialogue?tab=readme-ov-file}{github.com/google-research-datasets/dstc8-
schema-guided-dialogue}}to validate the generalizability of our approach and ensure its robustness across different datasets.

% \section{Conclusion}
\section{Acknowledgments}
This research was supported in part by Other Transaction award HR0011249XXX from the U.S. Defense Advanced Research Projects Agency (DARPA) Friction forAccountability in Conversational Transactions (FACT) program. This research project has benefited from the Microsoft Accelerate Foundation Models Research (AFMR) grant program through which leading foundation models hosted by Microsoft Azure along with access to Azure credits were provided to conduct the research.
\newpage
\bibliographystyle{IEEEbib}
\bibliography{refs}

\newpage
\onecolumn
\appendix
\section*{Appendix}

\section{Zero-shot Prompt Foramt}
\begin{center}
\begin{minipage}{0.9\linewidth}
\begin{verbatim}
<|begin_of_text|>
<|start_header_id|>system<|end_header_id|>
Capture entity values from the LAST UTTERANCE of the conversation.
FOCUS ONLY ON THE VALUES MENTIONED IN THE LAST UTTERANCE.
Format the output as a valid JSON object for each entity-value pair.
Format: {"state": {"_entity_":"_value_"}, "confidence": "X"}
Where X is the Confidence of the answer.

Fill the actual entity value into the placeholder encapsulated with underscores.
Put "```" as EOS token at the end of response.
Values that should be captured are:

In the DOMAIN of "train", the values that should be captured are:
 - "arriveby" that specifies what time the train should arrive
 - "leaveat" that specifies what time the train should leave
 - "day" that specifies what day the train should leave 
 (monday/tuesday/wednesday/thursday/friday/saturday/sunday)
 - "departure" that specifies the departure station
 - "destination" that specifies the destination station
 - "bookpeople" that specifies how many people the booking is for

Do not capture any other values!
If not specified, do not respond to that slot-value.

MAKE SURE TO SEPARATE EACH SLOT-VALUE PAIR, AND ALONG WITH EACH OF THEIR CONFIDENCE (0-1).
Format the output as:
```json
[
    {"state": {"_entity1_":"_value1_"}, "confidence": "_X1_"}, 
    {"state": {"_entity2_":"_value2_"}, "confidence": "_X2_"},
    {"state": {"_entity3_":"_value3_"}, "confidence": "_X3_"},
]```

Now complete the following example, AND PROVIDE CONFIDENCE THAT IT'S CORRECT:
input: <|eot_id|>
<|start_header_id|>user<|end_header_id|>
Customer: I need train reservations from norwich to cambridge
<|eot_id|>

<|start_header_id|>assistant<|end_header_id|>
***Output JSON format***
Output: ```json[
\end{verbatim}
\end{minipage}
\end{center}

\section{Few-shot Prompt Foramt}
\begin{center}
\begin{minipage}{0.9\linewidth}
\begin{verbatim}
<|begin_of_text|>
<|start_header_id|>system<|end_header_id|>
Capture entity values from the LAST UTTERANCE of the conversation.
FOCUS ONLY ON THE VALUES MENTIONED IN THE LAST UTTERANCE.
Format the output as a valid JSON object for each entity-value pair.
Format: {"state": {"_entity_":"_value_"}, "confidence": "X"}
Where X is the Confidence of the answer.

Fill the actual entity value into the placeholder encapsulated with underscores.
Put "```" as EOS token at the end of response.
Values that should be captured are:

In the DOMAIN of "train", the values that should be captured are:
 - "arriveby" that specifies what time the train should arrive
 - "leaveat" that specifies what time the train should leave
 - "day" that specifies what day the train should leave 
    (monday/tuesday/wednesday/thursday/friday/saturday/sunday)
 - "departure" that specifies the departure station
 - "destination" that specifies the destination station
 - "bookpeople" that specifies how many people the booking is for

Do not capture any other values!
If not specified, do not respond to that slot-value.
--------------------
---------------------Example 0:
context: Customer: I need to get a train from cambridge to Norwich.

state: ```json{{"state": {departure: cambridge}, "confidence": "0.9517114799962094"},
{"state": {destination: norwich}, "confidence": "0.9517114799962094"},
}```
---------------------Example 1:
context: Customer: I need a train from Cambridge to Norwich please.

state: ```json{{"state": {departure: cambridge}, "confidence": "0.909169572100446"},
{"state": {destination: norwich}, "confidence": "0.909169572100446"},
}```
--------------------
MAKE SURE TO SEPARATE EACH SLOT-VALUE PAIR, AND ALONG WITH EACH OF THEIR CONFIDENCE (0-1).
Format the output as:
```json
[
    {"state": {"_entity1_":"_value1_"}, "confidence": "_X1_"}, 
    {"state": {"_entity2_":"_value2_"}, "confidence": "_X2_"},
    {"state": {"_entity3_":"_value3_"}, "confidence": "_X3_"},
]```

Now complete the following example, AND PROVIDE CONFIDENCE THAT IT'S CORRECT:
input: <|eot_id|>
<|start_header_id|>user<|end_header_id|>
Customer: I need train reservations from norwich to cambridge
<|eot_id|>

<|start_header_id|>assistant<|end_header_id|>
***Output JSON format***
Output: ```json[
\end{verbatim}
\end{minipage}
\end{center}

\section{Self-Probing Prompt}
\subsection{Turn-level}
\begin{center}
\begin{minipage}{0.9\linewidth}
\begin{verbatim}
Conversation:
['Customer: I need train reservations from norwich to cambridge']
State:
{''departure': 'norwich', 'destination': 'cambridge', 'bookpeople': '2'}
How likely is the above state to be correct?
Analyze the state, provide 1 brief reason, and give confidence (0-1).
Format: confidence: "X", reason: "brief reason"
Think carefully and step by step.
Output:
\end{verbatim}
\end{minipage}
\end{center}

\subsection{Slot-level}
\begin{center}
\begin{minipage}{0.9\linewidth}
\begin{verbatim}
Conversation:
['Customer: I need train reservations from norwich to cambridge']
State:
{'arriveby': '-1'}
How likely is the above state to be correct?
Analyze the state, provide 1 brief reason, and give confidence (0-1).
Format: confidence: "X", reason: "brief reason"
Think carefully and step by step.
Output:
\end{verbatim}
\end{minipage}
\end{center}

\section{Ground Truth Confidence Score Prompt}
\begin{center}
\begin{minipage}{0.9\linewidth}
\begin{verbatim}
<|begin_of_text|>
<|start_header_id|>system<|end_header_id|>
You are a helpful AI assistant for evaluating the hardness of dialogue state tracking 
from last user utterance given dialogue history <|eot_id|>
<|start_header_id|>user<|end_header_id|>
How difficult would it be for a Language Model to predict the dialogue state from:
utterance: ['Customer: I need train reservations from norwich to cambridge']
given dialogue history
history:
['Customer: I need train reservations from norwich to cambridge']

Choose the level of hardness from (Easy/Medium/Hard).
Answer:
\end{verbatim}
\end{minipage}
\end{center}

\section{Prompt Variations}
\subsection{All Slot Description + All Slot Examples}
\begin{center}
\begin{minipage}{0.9\linewidth}
\begin{verbatim}
<|begin_of_text|>
<|start_header_id|>system<|end_header_id|>
Capture entity values from the LAST UTTERANCE of the conversation.
FOCUS ONLY ON THE VALUES MENTIONED IN THE LAST UTTERANCE.
Format the output as a valid JSON object for each entity-value pair.
Format: {"state": {"_entity_":"_value_"}, "confidence": "X"}
Where X is the Confidence of the answer.

Fill the actual entity value into the placeholder encapsulated with underscores.
Put "```" as EOS token at the end of response.
Values that should be captured are:

In the DOMAIN of "train", the values that should be captured are:
 - "arriveby" that specifies what time the train should arrive
 - "leaveat" that specifies what time the train should leave
 - "day" that specifies what day the train should leave 
    (monday/tuesday/wednesday/thursday/friday/saturday/sunday)
 - "departure" that specifies the departure station
 - "destination" that specifies the destination station
 - "bookpeople" that specifies how many people the booking is for

Do not capture any other values!
If not specified, do not respond to that slot-value.
--------------------
---------------------Example 0:
context: Customer: I need to get a train from cambridge to Norwich.

state: ```json{{"state": {departure: cambridge}, "confidence": "0.9517114799962094"},
{"state": {destination: norwich}, "confidence": "0.9517114799962094"},
}```
---------------------Example 1:
context: Customer: I need a train from Cambridge to Norwich please.

state: ```json{{"state": {departure: cambridge}, "confidence": "0.909169572100446"},
{"state": {destination: norwich}, "confidence": "0.909169572100446"},
}```
--------------------
MAKE SURE TO SEPARATE EACH SLOT-VALUE PAIR, AND ALONG WITH EACH OF THEIR CONFIDENCE (0-1).
Format the output as:
```json
[
    {"state": {"_entity1_":"_value1_"}, "confidence": "_X1_"}, 
    {"state": {"_entity2_":"_value2_"}, "confidence": "_X2_"},
    {"state": {"_entity3_":"_value3_"}, "confidence": "_X3_"},
]```

Now complete the following example, AND PROVIDE CONFIDENCE THAT IT'S CORRECT:
input: <|eot_id|>
<|start_header_id|>user<|end_header_id|>
Customer: I need train reservations from norwich to cambridge
<|eot_id|>

<|start_header_id|>assistant<|end_header_id|>
***Output JSON format***
Output: ```json[
\end{verbatim}
\end{minipage}
\end{center}

\subsection{All Slot Description + One Slot Examples}
\begin{center}
\begin{minipage}{0.9\linewidth}
\begin{verbatim}
<|begin_of_text|>
<|start_header_id|>system<|end_header_id|>
Capture entity values from the LAST UTTERANCE of the conversation.
FOCUS ONLY ON THE VALUES MENTIONED IN THE LAST UTTERANCE.
Format the output as a valid JSON object for each entity-value pair.
Format: {"state": {"_entity_":"_value_"}, "confidence": "X"}
Where X is the Confidence of the answer.

Fill the actual entity value into the placeholder encapsulated with underscores.
Put "```" as EOS token at the end of response.
Values that should be captured are:

In the DOMAIN of "train", the values that should be captured are:
 - "arriveby" that specifies what time the train should arrive
 - "leaveat" that specifies what time the train should leave
 - "day" that specifies what day the train should leave 
    (monday/tuesday/wednesday/thursday/friday/saturday/sunday)
 - "departure" that specifies the departure station
 - "destination" that specifies the destination station
 - "bookpeople" that specifies how many people the booking is for

Do not capture any other values!
If not specified, do not respond to that slot-value.
--------------------
---------------------Example 0:
context: Customer: I need to get a train from cambridge to Norwich.

state: ```json{{"state": {departure: cambridge}, "confidence": "0.9517114799962094"}
}```
---------------------Example 1:
context: Customer: I need a train from Cambridge to Norwich please.

state: ```json{{"state": {departure: cambridge}, "confidence": "0.909169572100446"}
}```
--------------------
MAKE SURE TO SEPARATE EACH SLOT-VALUE PAIR, AND ALONG WITH EACH OF THEIR CONFIDENCE (0-1).
Format the output as:
```json
[
    {"state": {"_entity1_":"_value1_"}, "confidence": "_X1_"}, 
    {"state": {"_entity2_":"_value2_"}, "confidence": "_X2_"},
    {"state": {"_entity3_":"_value3_"}, "confidence": "_X3_"},
]```

Now complete the following example, AND PROVIDE CONFIDENCE THAT IT'S CORRECT:
input: <|eot_id|>
<|start_header_id|>user<|end_header_id|>
Customer: I need train reservations from norwich to cambridge
<|eot_id|>

<|start_header_id|>assistant<|end_header_id|>
***Output JSON format***
Output: ```json[
\end{verbatim}
\end{minipage}
\end{center}

\subsection{One Slot Description + All Slot Examples}
\begin{center}
\begin{minipage}{0.9\linewidth}
\begin{verbatim}
<|begin_of_text|>
<|start_header_id|>system<|end_header_id|>
Capture entity values from the LAST UTTERANCE of the conversation.
FOCUS ONLY ON THE VALUES MENTIONED IN THE LAST UTTERANCE.
Format the output as a valid JSON object for each entity-value pair.
Format: {"state": {"_entity_":"_value_"}, "confidence": "X"}
Where X is the Confidence of the answer.

Fill the actual entity value into the placeholder encapsulated with underscores.
Put "```" as EOS token at the end of response.
Values that should be captured are:

In the DOMAIN of "train", the values that should be captured are:
 - "departure" that specifies the departure station

Do not capture any other values!
If not specified, do not respond to that slot-value.
--------------------
---------------------Example 0:
context: Customer: I need to get a train from cambridge to Norwich.

state: ```json{{"state": {departure: cambridge}, "confidence": "0.9517114799962094"},
{"state": {destination: norwich}, "confidence": "0.9517114799962094"},
}```
---------------------Example 1:
context: Customer: I need a train from Cambridge to Norwich please.

state: ```json{{"state": {departure: cambridge}, "confidence": "0.909169572100446"},
{"state": {destination: norwich}, "confidence": "0.909169572100446"},
}```
--------------------
MAKE SURE TO SEPARATE EACH SLOT-VALUE PAIR, AND ALONG WITH EACH OF THEIR CONFIDENCE (0-1).
Format the output as:
```json
[
    {"state": {"_entity1_":"_value1_"}, "confidence": "_X1_"}, 
    {"state": {"_entity2_":"_value2_"}, "confidence": "_X2_"},
    {"state": {"_entity3_":"_value3_"}, "confidence": "_X3_"},
]```

Now complete the following example, AND PROVIDE CONFIDENCE THAT IT'S CORRECT:
input: <|eot_id|>
<|start_header_id|>user<|end_header_id|>
Customer: I need train reservations from norwich to cambridge
<|eot_id|>

<|start_header_id|>assistant<|end_header_id|>
***Output JSON format***
Output: ```json[
\end{verbatim}
\end{minipage}
\end{center}

\subsection{One Slot Description + One Slot Examples}
\begin{center}
\begin{minipage}{0.9\linewidth}
\begin{verbatim}
<|begin_of_text|>
<|start_header_id|>system<|end_header_id|>
Capture entity values from the LAST UTTERANCE of the conversation.
FOCUS ONLY ON THE VALUES MENTIONED IN THE LAST UTTERANCE.
Format the output as a valid JSON object for each entity-value pair.
Format: {"state": {"_entity_":"_value_"}, "confidence": "X"}
Where X is the Confidence of the answer.

Fill the actual entity value into the placeholder encapsulated with underscores.
Put "```" as EOS token at the end of response.
Values that should be captured are:

In the DOMAIN of "train", the values that should be captured are:
 - "departure" that specifies the departure station

Do not capture any other values!
If not specified, do not respond to that slot-value.
--------------------
---------------------Example 0:
context: Customer: I need to get a train from cambridge to Norwich.

state: ```json{{"state": {departure: cambridge}, "confidence": "0.9517114799962094"}
}```
---------------------Example 1:
context: Customer: I need a train from Cambridge to Norwich please.

state: ```json{{"state": {departure: cambridge}, "confidence": "0.909169572100446"}
}```
--------------------
MAKE SURE TO SEPARATE EACH SLOT-VALUE PAIR, AND ALONG WITH EACH OF THEIR CONFIDENCE (0-1).
Format the output as:
```json
[
    {"state": {"_entity1_":"_value1_"}, "confidence": "_X1_"}, 
    {"state": {"_entity2_":"_value2_"}, "confidence": "_X2_"},
    {"state": {"_entity3_":"_value3_"}, "confidence": "_X3_"},
]```

Now complete the following example, AND PROVIDE CONFIDENCE THAT IT'S CORRECT:
input: <|eot_id|>
<|start_header_id|>user<|end_header_id|>
Customer: I need train reservations from norwich to cambridge
<|eot_id|>

<|start_header_id|>assistant<|end_header_id|>
***Output JSON format***
Output: ```json[
\end{verbatim}
\end{minipage}
\end{center}

\section{Domain Classification Prompt}
\begin{center}
\begin{minipage}{0.9\linewidth}
\begin{verbatim}
Determine which domain is considered in the following dialogue situation.
Choose one domain from this list:
 - restaurant
 - hotel
 - attraction
 - taxi
 - train
Answer with only one word, the selected domain from the list.
You have to always select the closest possible domain and never predict more than one word.
Consider the last domain mentioned, so focus mainly on the last utterance.

-------------------
Example1:
Customer: I need a cheap place to eat
Assistant: We have several not expensive places available. What food are you interested in?
Customer: Chinese food.

Domain: restaurant

-------

Example 2:
Customer: I also need a hotel in the north.
Assistant: Ok, can I offer you the Molly's place?
Customer: What is the address?

Domain: hotel

---------

Example 3:
Customer: What is the address?
Assistant: It's 123 Northfolk Road.
Customer: That's all. I also need a train from London.

Domain: train
""

Now complete the following example:
[history]
\end{verbatim}
\end{minipage}
\end{center}

\section{MultiWOZ 2.2}
\begin{table*}[t]
    \hspace*{-0.6in}
    \begin{footnotesize}
    \begin{minipage}{0.48\linewidth}
        \centering
        \begin{tabular}{|c|c|c|c|c|c|c|}
        \hline
        \multicolumn{7}{|c|}{\textbf{Zero-shot}} \\
        \hline
        Model & Self- & Confidence & \multicolumn{4}{|c|}{Results}\\
        \cline{4-7}
        & Probing & Estimation & JGA & Slot-F & AUC & ECE \\
        \hline
        Mistral & no & softmax & & & 0.609 &  0.368  \\
        \cline{3-3} \cline{6-7}
         -Nemo & & minicons & & & 0.525 &  0.244 \\
         \cline{3-3} \cline{6-7}
         & & verbalized & & & 0.595 & 0.411  \\
         \cline{3-3} \cline{6-7}
         & & combined & & & 0.662 & 0.017 \\
         \cline{2-3} \cline{6-7}
         
         & turn & softmax & & & 0.604 & 0.297\\
        \cline{3-3} \cline{6-7}
         & & minicons & & & 0..509 & 0.253 \\
        \cline{3-3} \cline{6-7}
         & & verbalized & 23.5 & 74.79 & 0.555 & 0.336\\
         \cline{3-3} \cline{6-7}
         & & combined & & & 0.617 & 0.013\\
         \cline{2-3} \cline{6-7}

         & slot & softmax & & & 0.597 & 0.358 \\
        \cline{3-3} \cline{6-7}
         & & minicons & & & 0.518 & 0.252 \\
        \cline{3-3} \cline{6-7}
         & & verbalized & & & 0.565 & 0.400 \\
         \cline{3-3} \cline{6-7}
         & & combined & & & 0.623 & 0.013 \\ 
        \hline
        
        Finetuned & no & softmax & & & 0.636 & 0.231\\
        \cline{3-3} \cline{6-7}
         Mistral & & minicons & & & 0.515 & 0.353\\
        \cline{3-3} \cline{6-7}
         & & verbalized & & & 0.492 & 0.255\\
         \cline{3-3} \cline{6-7}
         & & combined & & & 0.639 & 0.014\\
         \cline{2-3} \cline{6-7}
         
         & turn & softmax & & & 0.669 & 0.169 \\
        \cline{3-3} \cline{6-7}
         & & minicons & & & 0.498 & 0.362 \\
        \cline{3-3} \cline{6-7}
         & & verbalized & 42.7 & 86.19 & 0.438 & 0.189 \\
         \cline{3-3} \cline{6-7}
         & & combined & & & 0.704 & 0.042   \\
         \cline{2-3} \cline{6-7}
         
         & slot & softmax & & & 0.639 & 0.262\\
        \cline{3-3} \cline{6-7}
         & & minicons & & & 0.508 & 0.318\\
        \cline{3-3} \cline{6-7}
         & & verbalized & & & 0.503 & 0.292 \\
         \cline{3-3} \cline{6-7}
         & & combined & & & 0.653 & 0.026\\
         \cline{2-3} \cline{6-7}
        \hline
    \end{tabular}
    \end{minipage}
    \hfill
    \begin{minipage}{0.48\linewidth}
        \centering
        \begin{tabular}{|c|c|c|c|c|c|c|c|}
        \hline
        \multicolumn{7}{|c|}{\textbf{Few-shot}} \\
        \hline
        Model& Self- & Confidence & \multicolumn{4}{|c|}{Results}\\
        \cline{4-7}
        & Probing & Estimation & JGA & Slot-F & AUC & ECE \\
        \hline
        Mistral  & no & softmax & & & 0.583 & 0.326\\
        \cline{3-3} \cline{6-7}
         -Nemo & & minicons & & & 0.532 & 0.269 \\
        \cline{3-3} \cline{6-7}
         & & verbalized & & & 0.515 & 0.355\\
         \cline{3-3} \cline{6-7}
         & & combined & & & 0.586 & 0.022\\
         \cline{2-3} \cline{6-7} 
         
         & turn & softmax & & & 0.563 & 0.256\\
        \cline{3-3} \cline{6-7}
         & & minicons & & & 0.528 & 0.285\\
        \cline{3-3} \cline{6-7}
         & & verbalized & 24.26 & 75.91& 0.500 & 0.299 \\
         \cline{3-3} \cline{6-7}
         & & combined & & & 0.576 & 0.042 \\
         \cline{2-3} \cline{6-7}
         
         & slot & softmax & & & 0.611 & 0.327\\
        \cline{3-3} \cline{6-7}
         & & minicons & & & 0.504 & 0.268\\
        \cline{3-3} \cline{6-7}
         & & verbalized & & & 0.543 & 0.371 \\
         \cline{3-3} \cline{6-7}
         & & combined & & & 0.620 & 0.005\\
         \cline{2-3} \cline{6-7}
        \hline
        
        Finetuned & no & softmax &  & &  0.724 & 0.196 \\
        \cline{3-3} \cline{6-7}
         Mistral & & minicons & & & 0.496 & 0.346 \\
        \cline{3-3} \cline{6-7}
         & & verbalized & & & 0.484 & 0.249\\ 
         \cline{3-3} \cline{6-7}
         & & combined & & & 0.737 & 0.053 \\
         \cline{2-3} \cline{6-7}
         
         & turn & softmax &  & & 0.665 & 0.185\\
        \cline{3-3} \cline{6-7}
         & & minicons & & & 0.499 & 0.342\\
        \cline{3-3} \cline{6-7}
         & & verbalized & 42.0 & 85.60 & 0.428 & 0.206\\
         \cline{3-3} \cline{6-7}
         & & combined & & & 0.700 & 0.033\\
        \cline{2-3} \cline{6-7} 
         
         & slot & softmax & & & 0.678 & 0.273\\
        \cline{3-3} \cline{6-7}
         & & minicons & & & 0.493 & 0.292\\
        \cline{3-3} \cline{6-7}
         & & verbalized & & & 0.552 & 0.309 \\
         \cline{3-3} \cline{6-7}
         & & combined & & & 0.704 & 0.030\\
         \hline
        
    \end{tabular}
    \end{minipage}
    \end{footnotesize}
    \caption{Experimental zero-shot (left) an few-shot (right) results for dialogue state tracking performance (JGA and Slot-F) and confidence estimation quality (AUC and ECE).}
    \label{tab:alternative2}

\vspace{-0.1in}
\end{table*}

\end{document}